\documentclass[letterpaper, 10 pt, conference]{ieeeconf}  

\IEEEoverridecommandlockouts                              

\usepackage{graphics} 
\usepackage{epsfig} 
\usepackage{times} 
\usepackage{amsmath} 
\usepackage{amssymb}  

\usepackage{color,xcolor}
\usepackage{epsfig}
\usepackage{graphicx}

\usepackage{adjustbox}
\usepackage{array}
\usepackage{booktabs}
\usepackage{colortbl}
\usepackage{float,wrapfig}
\usepackage{hhline}
\usepackage{multirow}
\usepackage{subcaption} 
\usepackage[font={small}]{caption}

\usepackage{amsmath,amsfonts,amssymb}
\usepackage{bm}
\usepackage{nicefrac}
\usepackage{microtype}

\usepackage{changepage}
\usepackage{extramarks}
\usepackage{fancyhdr}
\usepackage{lastpage}
\usepackage{setspace}
\usepackage{soul}
\usepackage{xspace}

\usepackage{url}

\usepackage{enumerate}

\newcommand{\fig}[1]{Figure~\ref{#1}}

\newcommand{\ignorethis}[1]{}

\makeatletter
\DeclareRobustCommand\onedot{\futurelet\@let@token\@onedot}
\def\@onedot{\ifx\@let@token.\else.\null\fi\xspace}

\def\eg{e.g\onedot} 
\def\ie{i.e\onedot}

\def\etal{et al\onedot}
\makeatother

\definecolor{MyDarkBlue}{rgb}{0,0.08,1}
\definecolor{MyDarkGreen}{rgb}{0.02,0.6,0.02}
\definecolor{MyDarkRed}{rgb}{0.8,0.02,0.02}
\definecolor{MyDarkOrange}{rgb}{0.40,0.2,0.02}
\definecolor{MyPurple}{RGB}{111,0,255}
\definecolor{MyRed}{rgb}{1.0,0.0,0.0}
\definecolor{MyGold}{rgb}{0.75,0.6,0.12}
\definecolor{MyDarkgray}{rgb}{0.66, 0.66, 0.66}

\newcommand{\atcvam}{A3C (V+A+M+Mapper)\xspace}

\usepackage{paralist}

\title{\LARGE {\bf
Look, Listen, and Act: Towards Audio-Visual Embodied Navigation}
}

\author{Chuang Gan*$^{1}$, Yiwei Zhang*$^{2}$, Jiajun Wu$^{3}$,  Boqing Gong$^{4}$, Joshua B. Tenenbaum$^{3}$
\thanks{*indicates equal contributions. $^{1}$MIT-IBM Watson AI Lab, $^{2}$Tsinghua University, $^{3}$Massachusetts Institute of Technology, $^{4}$Google. Project page: \protect\url{http://avn.csail.mit.edu}}}

\begin{document}

\maketitle
\thispagestyle{empty}
\pagestyle{empty}


\begin{abstract}
 A crucial ability of mobile intelligent agents is to integrate the evidence from multiple sensory inputs in an environment and to make a sequence of actions to reach their goals. In this paper, we attempt to approach the problem of Audio-Visual Embodied Navigation, the task of planning the shortest path from a random starting location in a scene to the sound source in an indoor environment, given only raw egocentric visual and audio sensory data. To accomplish this task, the agent is required to learn from various modalities, \ie, relating the audio signal to the visual environment. Here we describe an approach to audio-visual embodied navigation that takes advantage of both visual and audio pieces of evidence. Our solution is based on three key ideas: a visual perception mapper module that constructs its spatial memory of the environment, a sound perception module that infers the relative location of the sound source from the agent, and a dynamic path planner that plans a sequence of actions based on the audio-visual observations and the spatial memory of the environment to navigate toward the goal. Experimental results on a newly collected Visual-Audio-Room dataset using the simulated multi-modal environment demonstrate the effectiveness of our approach over several competitive baselines. 
\end{abstract}

\section{Introduction}
We humans perceive and navigate through the world by an integration of multiple sensory inputs, including but not limited to acoustic and visual representations. Extensive evidence from cognitive
psychology and neuroscience suggests that humans, even in the young-child ages, are remarkably capable of capturing the correspondences between different signal modalities to arrive at a thorough understanding of natural phenomena~\cite{mcgurk1976hearing,kording2007causal}. 

\begin{figure}[t]
   \centering
   \includegraphics[width = 1\linewidth]{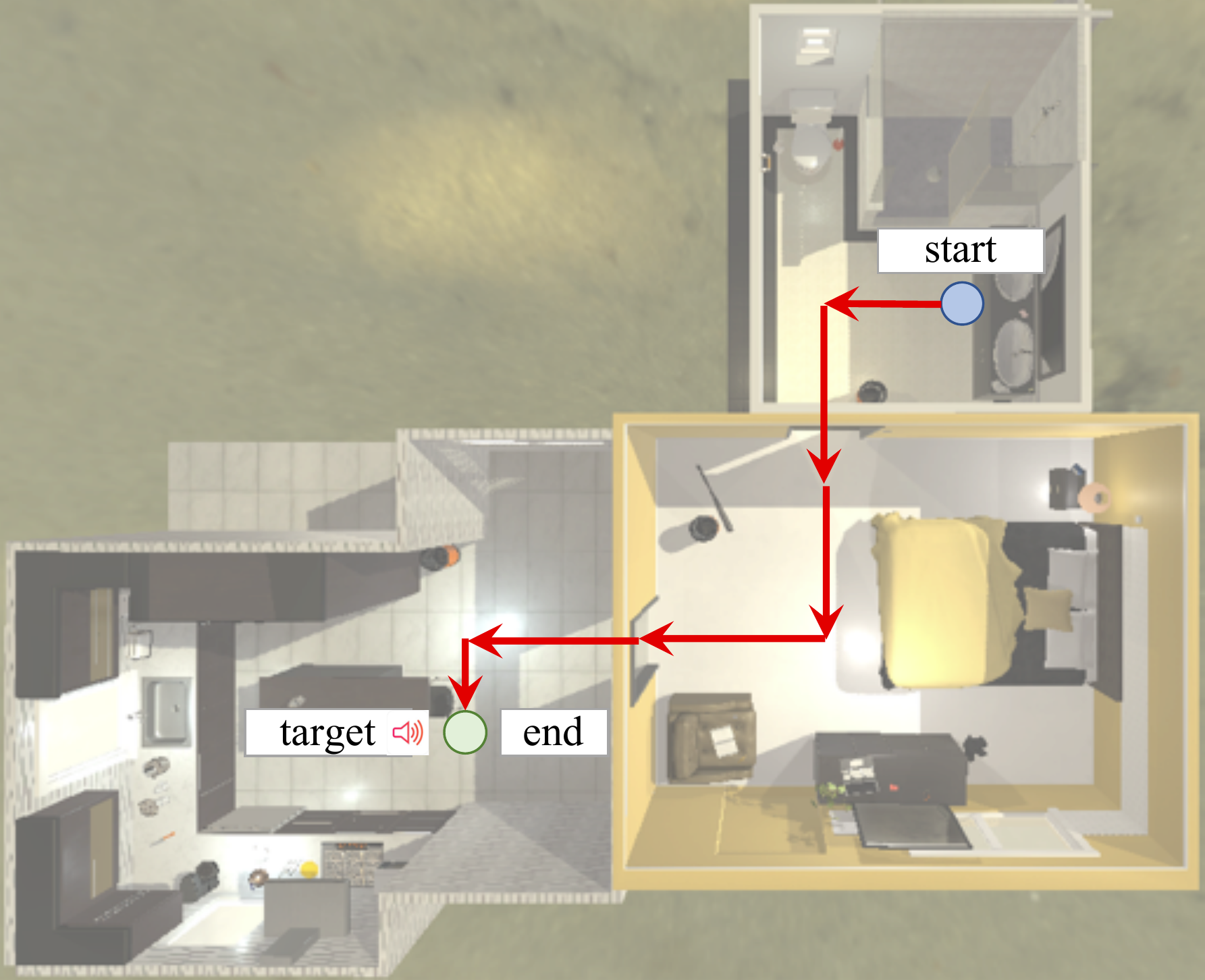}
   \caption{Illustration of the proposed Audio-Visual Embodied Navigation.  Given a sound source, the agent perceives and relates  it with the visual environment so as to  navigate towards the goal.}
   \vspace{-18pt}
   \label{fig:teaser}
\end{figure}

In this work, we aim to build a machine model to achieve such human competency on audio-visual scene analysis. Particularly, we focus on an under-explored problem: how to teach an intelligent agent, equipped with a camera and a microphone, to interact with environments and navigate to the sound source. This task is more complex and challenging than understanding static images. We call it \textit{audio-visual embodied navigation}. The agent has to sequentially make decisions from multiple sensory observations in order to navigate through an environment. 

\fig{fig:teaser} illustrates an example of the audio-visual embodied navigation problem. An agent starts randomly from an indoor environment and provided a sound impulse (\eg., a phone is ringing). The agent then needs to figure out the shortest trajectory to navigate from its starting location to the source of the sound (the goal location). To accomplish this task, the agent must relate the sound to the visual environment and meanwhile map from the raw sensory inputs to the actions of path planing. This task is very challenging, because it essentially asks the agent to reverse-engineer the causal effect of the signals it perceives. It has to not only understand the visual surrounding but also reason upon it for the potential sound source. If the sound source is in a different room, the agent additionally has to learn how to move to another room. Progress on this task will facilitate many real-world applications in home robots. For example, the sound-source-seeking ability can empower the robots to help humans find their cellphones or turn off faucets.


We tackle this problem by drawing insights from how humans complete this task. Imagine how a user seeks the sound source in a novel environment? S/he first gets a sense about the location, the surrounding, as well as an estimate about the distance and direction to sound source. As s/he moves around, the room layout and her/his memory about the sound and the room also play vital roles when s/he makes decisions for future actions. To this end, in this paper, we design an audio-visual network to let the machine agent imitate the navigation mechanism we humans use in real life. We systematically study two settings: explore-and-act and non-exploration.  In the first setting, the agent is allowed to explore the environment up a certain number of steps before we alarm the sound;  Thus it can build an incomplete spatial memory of the room from visual observations during the exploration and can refer back to this internal representation in resolving the subsequent navigation tasks. In the second setting, the agent has to build a dynamic spatial map during the navigation.  When the agent is presented a task of searching for the sound source, akin to humans, the agent must first infer the relative position of the sound source and then makes sequential actions to move towards the source based on its spatial memory and visual-audio observations. It can also update its spatial memory at each time step. 

To sum up, our work makes the following contributions:
 \begin{compactitem}
	\item We build a multi-modal navigation environment to facilitate research on audio-visual embodied navigation. The environment contains fairly complex apartments and has an integrated sound module that observes certain physical laws.
	
	\item We present a Visual-Audio Room (VAR) benchmark to systematically evaluate the performances of multi-modal navigation agents.
	
	\item We propose a framework for audio-visual embodied navigation and contrast it to several competitive baselines. Experiments verify the effectiveness of our approach and meanwhile reveal the key challenges in this task, shedding lights on future work.
\end{compactitem}
 
 
\section{Related Work}
\label{sec:related}

\subsection{Target-driven Navigation}
Early work solves the navigation task by building a map of the scene using SLAM and then planing the path in this map~\cite{thrun2005probabilistic}. More recently, deep learning--based methods have been used to plan actions directly from raw sensory data. Zhu~\etal\cite{zhu2017target} investigate a reactive deep RL based navigation approach to find the picture of the target object in a discretized 3D environment. Gupta~\etal\cite{gupta2017cognitive} learn to navigate via a mapper and planner. Sadeghi~\etal\cite{sadeghi2016cad2rl} study a RL approach that can solely use simulated data to teach an agent to fly in real. Mirowski~\etal\cite{mirowski2016learning} improve the navigation results  by jointly training with auxiliary tasks such as loop closure detection and depth estimations from RGB. Brahmbhatt~\etal\cite{brahmbhatt2017deepnav} explore a CNN architecture to navigate large cities with street-view images. Wu~\etal\cite{wu2016training} combine deep RL with curriculum learning in a first-person shooting game setting.   Yang~\etal\cite{yang2018visual} propose to improve visual navigation with semantic priors. McLeod~\etal\cite{mcleod2019navigating} leverage past experience to improve future robot navigation in dynamically unknown environments. Katyal~\etal\cite{katyal2019uncertainty} use generating networks to predict occupancy map representations for future robot motions. Mousavian~\etal\cite{mousavian2018visual} demonstrate the use of semantic segmentation and detection masks for target driven visual navigation. The work by Savinov~\etal~\cite{savinov2018semi} is inspirational to ours in terms of the use of memory in navigation. Unlike their topological graph memory, we use a key-value structure to better capture the incomplete nature of our agent’s internal knowledge about the environment and also introduce a dynamic spatial memory for the non-exploration setting.

There is also considerable work on vision-language based embodied navigation~\cite{fried2018speaker,yu2018interactive,mei2016listen,anderson2018vision}, which map directly from language instructions and visual observations to actions. In contrast to all these approaches, the goal of embodied audio-visual navigation task is to find the sound source, thus require the agent to integrate the visual and acoustic cues to plan a sequence of actions. Concurrent to our work, Chen \etal~\cite{chen2019audio} also poses the task of audio-visual embodied navigation on the Habitat platform~\cite{savva2019habitat}. 

\subsection{Sound Localization}

The research problem of sound localization, which aims to identify which regions in a video make the sound, has been studied for decades. Existing approaches fall into two broad categories: computation-based approaches and learning-based approaches. Early work measure the correlations between pixels and sounds using a Gaussian process model~\cite{hershey2000audio}, subspace methods~\cite{fisher2001learning}, canonical correlation analysis~\cite{kidron2005pixels}, hand-crafted motion~\cite{barzelay2007harmony}, and segmentation~\cite{izadinia2013multimodal}. More recently, researchers have proposed to train a deep neural network to see and hear many unlabeled videos to localize the objects that sound~\cite{arandjelovic2017objects,senocak2018learning,ephrat2018looking,afouras2018conversation,owens2018audio,gao2018learning, zhao2018sound,zhao2019sound, rouditchenko2019self,gan2019self}. Such methods address the task of localizing the regions on videos. In contrast, our goal is to find the sound source that might also be non-line-of-sight in the virtual environment. Our work is also related to acoustic based approaches for the sound source localization~\cite{an2018reflection,van2004optimum,zunino2015seeing}. They typically require special devices (\eg microphone arrays) to record the sounds.  Our work goes beyond that, since the mobile agent needs to further connect the result of sound localization with the visual environment for the path planing.

\section{Environment}
\label{sec:environment}

We build a multi-modal virtual environment on top of the AI2-THOR platform~\cite{kolve2017ai2}, which contains a set of scenes built by using the Unity Game engine. We further incorporate an spatial audio software development kit, Resonance Audio API~\cite{gorzel2019efficient}, into the Unity game engine to support the audio-visual embodied navigation task. An agent equipped with a camera and a microphone can then leverage two different sensing modalities (egocentric RGB images and sound) to perceive and navigate through the scenes. 

\subsection{Scene} 

The AI2-THOR platform provides near photo-realistic indoor scenes. It contains 120 scenes in four categories: \emph{kitchens},  \emph{bedrooms}, \emph{living rooms}, and \emph{bathrooms}. Each room category contains 30 rooms with diverse visual appearances. As the rooms in AI2-THOR are isolated from each other, they are not challenging enough for the area-goal navigation tasks. As shown in Figure~\ref{fig:teaser},  we instead manually concatenate several rooms into multi-room apartments. We were able to build seven apartments in total for this work. Similarly to Zhu~\etal\cite{zhu2017target}, we also discretize the apartments into grid worlds for the ease of quantitative evaluation. Particularly, each grid square is $0.5\times 0.5$ square meters. In total, there are about 150 to 200 grid cells for each apartment.

\subsection{Sound} 
The acoustic engine is implemented using the Resonance Audio API. This spatial audio SDK can create high fidelity spatial sound in the virtual environment and support acoustic ray-tracing according to the geometry of object and house. The Resonance Audio API also supports the simulation of how sound waves interact with objects in various geometry and diverse materials by mapping visual materials to acoustic materials with realistic frequency-dependent sound reverberation properties. How sound waves interact with human ears can also be simulated accurately based on interaural time differences, interaural level differences, and the spectral effects. The scenes equipped with the Resonance Audio API create the illusion that sounds come from specific locations in the virtual world. Thus we can use this module to provide the agent with stereo sound frames. In our experimental setting, we mainly consider the continuous sound sources (\eg, ring tone and alarm alert).

\begin{figure*}[t]
   \centering
   \includegraphics[width=1.0\linewidth]{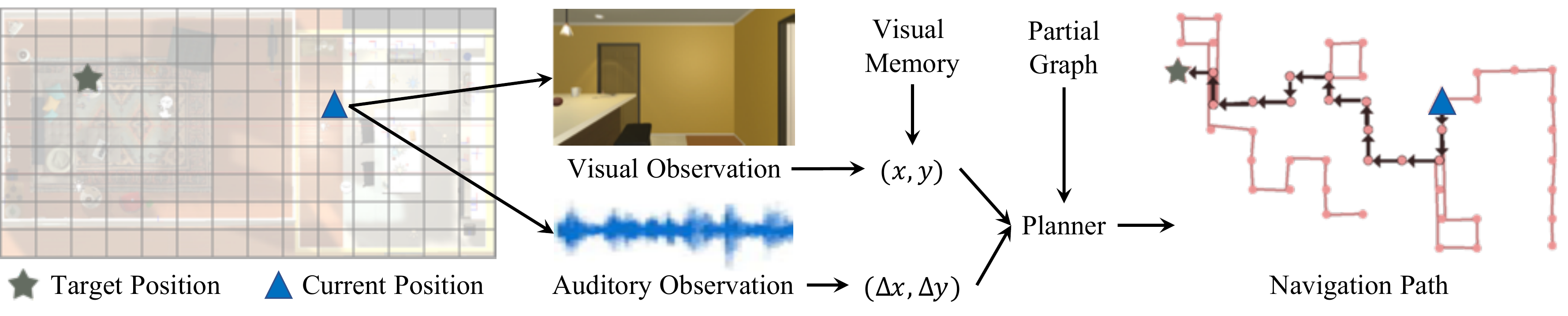}
   \vspace{-17pt}
    \caption{During navigation, the agent uses a visual perception module to localize itself and a sound perception module to estimate the sound source location. Finally, the agent connects the estimation with the partial graph built during exploration, and plans a path to reach the goal.} 
    \vspace{-12pt}
   \label{fig:navigation}
\end{figure*}



\subsection{Problem Setup}


We consider two problem setups in the visual-audio embodied navigation task: explore-and-act and non-exploration for seeking the sound sources. In the first setting, we let the agent conduct two stages of interactions with an environment, in spirit similar to the exploration setup  in Savinov~\etal\cite{savinov2018semi}: exploration and goal-directed navigation. During the exploration stage, the agent randomly walks around the environment until its trajectory's length reaches a given budget. This exploration experience can be utilized to build an internal world model of the environment for subsequent goal-directed navigation tasks. In the second setting, the agent has to simultaneously build a spatial memory of the environment while it navigates towards the goal.

In the testing phase, we randomly select a starting position for the agent and a target position for the sound source. At each time step, the agent takes an egocentric visual observation $o_v$ and sound observation $o_s$ about the environment and then draws an action from a set $A$. The action set $A$ comprises the following: \textit{Move Backward}, \textit{Move Forward}, \textit{Rotate Left}, \textit{Rotate Right}, and \textit{Stop}.
\section{Approach}
\label{sec:approach}

In this section, we introduce our algorithm for the audio-visual embodied navigation. As shown in \fig{fig:navigation}, it consists of three components: a visual perception mapper, an audio perception module, and a dynamic path planner. 

In an attempt to tackle the visual-audio embodied navigation task, the agent can either use the visual perception mapper to retrieve from its spatial memory constructed during the exploration stage or use the occupancy map estimated during navigation to build a \emph{partial} 2D spatial map (a graph with edges) of the environment. After that, the agent leverages the sound perception module to estimate the direction and distance of the sound source from its current location. Finally, the agent plans to find the shortest possible path to reach the goal based on the inference results from the visual and sound perception modules and the \emph{partial} spatial map. In the middle of the navigation, the dynamic planner can also update the the estimated goal location as well as the agent's navigation model based on any new visual and audio observations. 


\begin{figure*}[t]
   \centering
   \includegraphics[width=0.9\linewidth]{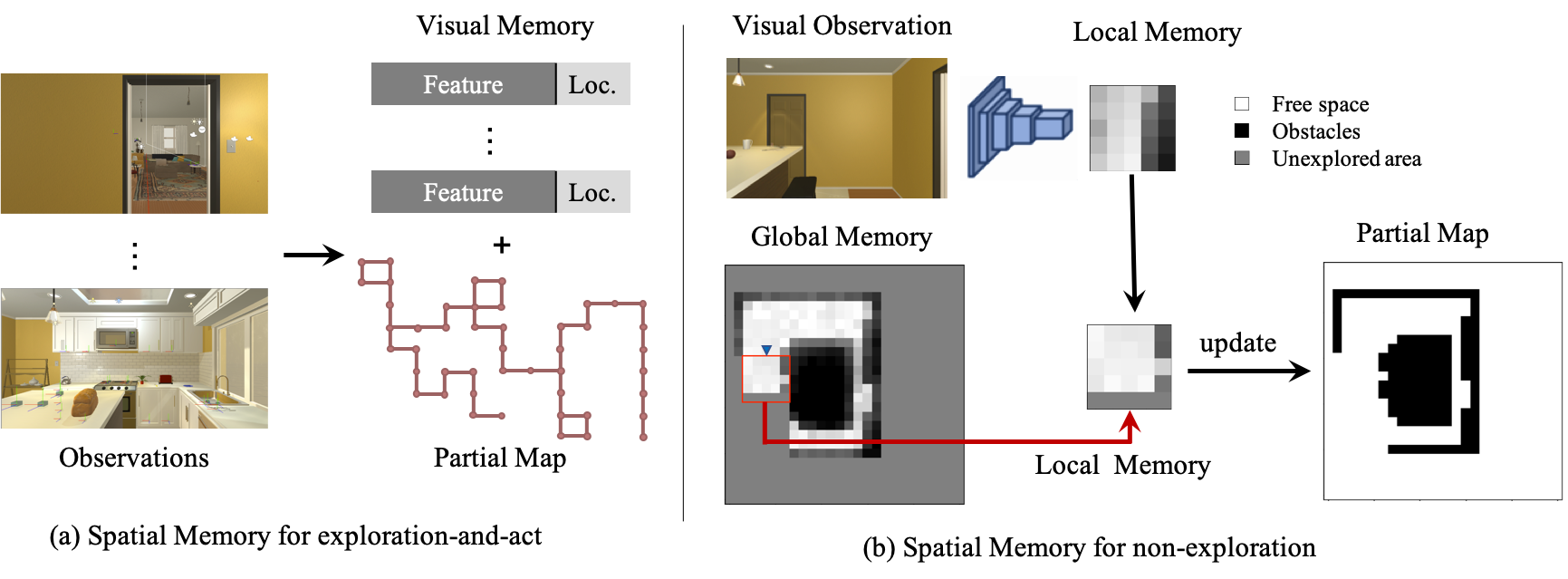}
   \vspace{-8pt}
   \caption{The spatial memory used for exploration-and-act and non-exploration based navigation.}
   \vspace{-15pt}
   \label{fig:explore}
\end{figure*}

\subsection{Visual Perception Mapper}

\paragraph{Explore-and-Act Visual Mapper.}
The exploration based visual mapper consists of a spatial memory and a non-parametric retrieval model. During the exploration, we use a key-value based spatial memory network~\cite{miller2016key} to encode the environment visited by the agent. Visual observation is stored in the key part, while the meta-data (\eg, location coordinates and actions taken by the agent) is stored in the value part. Concretely, we forward each first-person view RGB image through an ImageNet-pretrained ConvNet model, ResNet~\cite{he2016deep}, to extract a visual feature vector, followed by $\ell_2$ normalization. Each feature vector of the visual observation and the corresponding coordinates and orientation of the agent form a key-value pair (\fig{fig:explore} (a)). 

In the goal-oriented navigation stage, we use the same feature extraction pipeline to obtain a query feature vector extracted from the agent's first-person view. A non-parametric retrieval step is then conducted over the keys of the spatial memory. The top three memory slots that are closest to the query under the cosine distance are returned. We average the stored coordinates to estimate the location of the agent. Note that, in our experiments, we enforce the starting location of the agent has no overlap with the explored positions, so the localization of the agent in the early navigation stage has significant amount of uncertainty. 

\paragraph{Non-Exploration Visual Mapper.}
Since the random exploration stage might not always be allowed in real applications, we further introduce a visual perception mapper that can construct a dynamic spatial memory on the fly while the agent navigates through an environment. We construct the spatial memory as a top-down 2D occupancy map  $M \in R^{N \times N}$ about the grid world environment (\fig{fig:explore} (b)). Each entry of the memory represents a grid cell in the environment, taking values in the range [0, 1] and indicating the agent's belief about whether the grid cell can be traversed. We initialize the value to 0.5 and the starting position of the agent at the center of the memory $M$. Since the agent only has access to the visual observations of the nearby area, the visual perception mapper produces a local occupancy map in front of the agent at each time step. The agent then updates the local occupancy map using Bayes’ rule~\cite{moravec1989sensor}. 

In order to infer if a cell is tranversable or not, we train a convolutional neural network to map from RGB images to the agent's believes. We collect training data by utilizing RGB images and groundtruth 2D occupancy maps in two training apartments. We feed a first-person RGB image into the network to predict the free space probability in a small window size (5 $\times$ 5 grids in our experiments).  

\subsection{Sound Perception Module}
Motivated by the fact that humans can mentally infer the sound locations, here we introduce a sound perception module used for estimating the coordinates of the goal, \ie, the location of the sound source in our environment. 
 
To make the network easy to learn, we use relative locations to estimate the coordinates of the goal. We define the relative location in a grid world as follows. The relative location of the audio source from the agent’s perspective, denoted by $(x, y)$, represents that the audio source is located $x$ meters to the right of the agent and $y$ meters in front of the agent. The relative position can thus be calculated by jointly considering the orientation of the agent and the absolute coordinates of the sound source and the agent. In the absolute coordinate system, the east axis is the positive $x$ axis and the north axis is the positive $y$ axis. Assuming that the absolute coordinates of the sound source and the agent in the grid world are (16, 18) and (4, 8), respectively, the relative location of the sound source is then (12, 10) if the orientation of the agent is towards the north. If the agent faces the west, the relative location of the sound source becomes (10, -12). 

We collect training data by using all the recorded audio clips in two training apartments.  To pre-process the stereo sound, we first re-sample the sound to 16 HZ and then convert it into a spectrogram through Short-Time Fourier Transform (STFT). We feed the spectrogram into a five-layer convolutional network. The training objective is to estimate the relative sound location with respect to the agent. Specifically, we train the sound perception module with the mean square error (MSE) loss in a supervised way. The absolute coordinate of the target can be calculated from the relative location and the agent's coordinates. The sound perception module also contains noises, since the layout and the surface materials of the room influence the sound. We also train a separate sound classification model to decide if the agent reaches the goal; if so, it will take a \textit{stop} action.
 
\subsection{Dynamic Path Planner}
The agent employs a dynamic path planner to navigate in the environment upon observing changes and meanwhile updating the memory as a result of those actions.

For the explore-and-act setting, the agent can construct a \emph{partial} graph $G = (V, E)$ of the room based on a sequence of the agent's exploration trajectory ($x_0$, $x_1$, $\cdots$, $x_n$) and actions ($a_1$, $a_2$, $\cdots$, $a_n$). Denote by $V$ and $E$ the nodes and edges, respectively. Each node $v_i \in V$ in the graph stores a location (coordinates) in the environment. The edge exists between two nodes $v_i$ and $v_j$ if the nodes correspond to consecutive time steps.

For the non-exploration based approach, the agent can use the memory to construct a \emph{partial} graph of the environment. According to the spatial memory, each grid can be divided into three types: free spaces, obstacles, and unexplored areas. The \emph{partial} graph $G = (V, E)$  is generated by assuming the unexplored and unoccupied grids are transversable, converting the traversable grids into nodes, and connecting adjacent nodes via edges. When a new obstacle is detected, the corresponding node and edge(s) will be deleted. 

The dynamic planner outputs a path (sequence of actions) that transits the agent to the desired target location based on the graph of the environment and the estimated target coordinates.  At each time step $t$, the agent tries to find a shortest path from the graph node of the agent itself $x^t_{agent}$ to the graph node $x^t_{graph}$ that is nearest to the estimated coordinates of the sound source $x^t_{target}$. We use the Dijkstra’s algorithm to solve this problem. Specifically, we can use this algorithm to obtain a shortest path denoted as $p_{min}^t = \{a_0^t, a_1^t, \cdot , a_m^t\}$. We select $a^{t}_{0}$ as the action to take at time $t$.

\begin{figure*}[t]
   \centering
   \includegraphics[width=\linewidth]{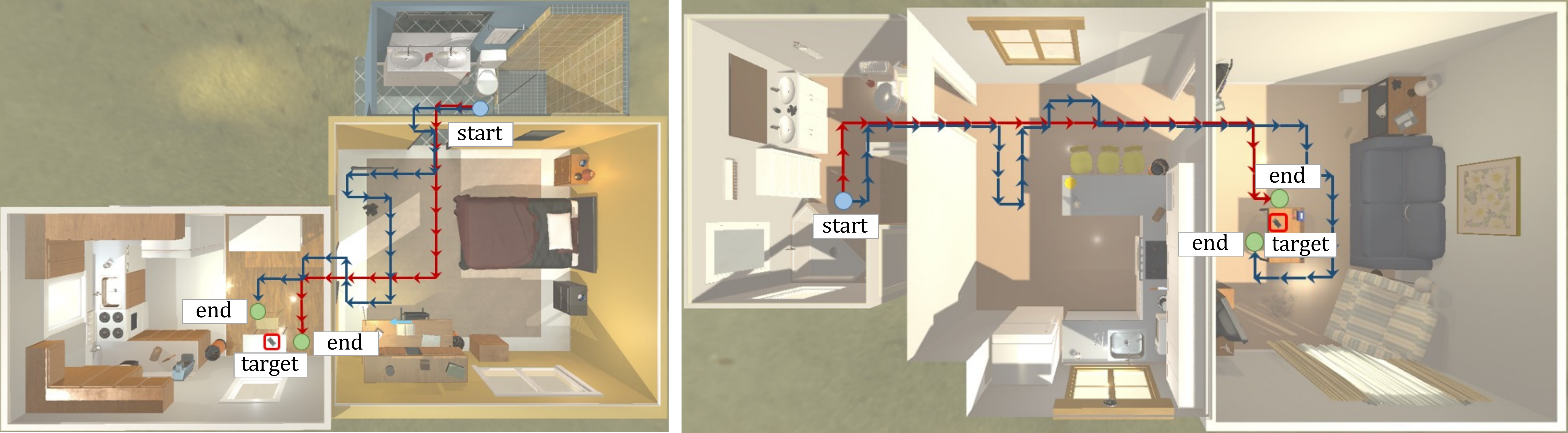}
    \caption{
    Trajectories of ours (red) against those of \atcvam (blue). Our model finds more efficient routes to the targets.}
    \label{fig:trajectory}
     \vspace{-15pt}
\end{figure*}

\section{Experiment}
\label{sec:experiment}

In this section, we first describe the dataset and environment we created for the sound-source-seeking audio-visual embodied navigation task, followed by demonstrating the benefits of our method against several competing baselines. 

\subsection{The Visual-Audio Room Dataset} 

In order to systematically evaluate the multi-modal navigation performances and facilitate future study on this research direction, we collect a new Visual-Audio Room (VAR) benchmark. Our ultimate goal is to enable robots to navigate in indoor environments by fusing visual and audio observations. As it is very challenging to conduct thorough and controlled experiment with physical robots, we instead use a 3D simulated environments, built on the top of AI-Thor platform~\cite{zhu2017target}. There are seven apartments used in the experiment. We split them to two apartments for training and five apartments for testing. We consider three recorded audio categories: ring tone, alert alarm, and clock clicking as the sound source in the work. To generate navigation data, we pre-extract the first person view RGB images in four orientations over all the grids of the seven apartments. For the training apartments, we randomly pick up ten locations to put the sound source and then record the sound the agent hears in four orientations of all locations in the room. For the five testing apartments, we pick five grid locations to put the sound source and also record the sound the agent hears in all locations. In total, we collect 3,728 RGB images, and record 75,720 audio clips. 

\subsection{Experiment Setup}
\noindent\textbf{Setup.} We aim to test the agent's ability to navigate in a new environment. During evaluation, for each of the five possible locations of the sound source in each room, we randomly pick 20 locations and orientations of the agent. Therefore, we conduct 100 testing episodes for each sound. We set the time duration of each sound at each time step as 1 second. The number of random walk steps in the explore-and-act setting are set as 400 . 

\noindent\textbf{Evaluation metrics.} We use two metrics to evaluate the models: success rates and Success weighted by Path Length (SPL)~\cite{anderson2018evaluation}.  
\textit{SPL} 
measures the agent’s navigation performance across all the testing scenarios as $\frac{1}{N} \sum_{i=1}^N  S_{i} \frac{l_i}{\max(p_i,l_i)}$, where $N$ denotes the number of testing episodes, $S_i$ is a binary indicator of success in the testing episode $i$, $P_i$ represents path length, and $l_i$ indicates the shortest distance to the goal.

\begin{table*}[t]
\centering
\small
\newcommand{\tabincell}[2]{\begin{tabular}{@{}#1@{}}#2\end{tabular}} 
\begin{tabular}{lccccccc}
\toprule
Approach& Sound &Room 1 & Room 2 & Room 3 &  Room 4 & Room 5 & Average               \\ 
\midrule
Random Search &   \tabincell{l}{Ring \\ Alarm \\ Clock}&  \tabincell{c}{6/3.5 \\ 5/3.3 \\8/5.2} &\tabincell{c}{10/6.8 \\ 9/5.7 \\9/6.6}  &\tabincell{c}{8/4.6 \\ 7/4.2 \\6/3.4}  &\tabincell{c}{5/2.6 \\ 7/4.1 \\4/2.1}  &\tabincell{c}{8/5.1 \\ 7/3.9 \\7/4.4}  &\tabincell{c}{7.4/4.5 \\ 7/4.2 \\6.8/4.3}  \\
\midrule
Greedy Search (A)   & \tabincell{l}{Ring \\ Alarm \\ Clock} &\tabincell{c}{12/10.5 \\ 9/8.2 \\13/12.0}  &\tabincell{c}{15/13.5 \\ 13/11.2 \\15/13.4}  &\tabincell{c}{13/11.8 \\ 10/8.4 \\12/11.0}  &\tabincell{c}{10/8.2 \\ 8/7.1 \\12/9.6}  &\tabincell{c}{12/11.4 \\ 11/10.2 \\11/10.1}  &\tabincell{c}{12.4/11.1 \\ 10.2/9.0 \\12.6/11.2}           \\
\midrule
  A3C (V)    &   \tabincell{l}{Ring \\ Alarm \\ Clock}&  \tabincell{c}{5/4.7 \\ 7/6.0 \\7/6.1}  &\tabincell{c}{12/11.5 \\ 11/9.3 \\8/7.5}  &\tabincell{c}{10/9.1 \\ 9/8.9 \\9/7.8}  &\tabincell{c}{8/7.8 \\ 7/6.2 \\9/8.6}  &\tabincell{c}{10/8.9 \\ 8/7.5 \\11/9.8}  &\tabincell{c}{9/8.4 \\ 8.4/7.6 \\8.8/8.0}  \\
\midrule
  A3C (V+A)    &   \tabincell{l}{Ring \\ Alarm \\ Clock}&  \tabincell{c}{18/15.7 \\ 19/15.9 \\18/13.0}  &\tabincell{c}{25/18.5 \\ 22/17.1 \\27/23.6}  &\tabincell{c}{15/11.7 \\ 16/14.5 \\18/13.8}  &\tabincell{c}{14/11.5 \\ 14/12.5 \\16/13.8}  &\tabincell{c}{20/14.9 \\ 19/14.6 \\22/16.1}  &\tabincell{c}{18.4/14.5 \\ 18/14.9 \\20.2/16.1}  \\
\midrule
   A3C (V+A+M)   &   \tabincell{l}{Ring \\ Alarm \\ Clock}&  \tabincell{c}{27/18.4 \\ 24/16.5 \\29/20.3}  &\tabincell{c}{36/25.7 \\ 37/25.0 \\34/25.6}  &\tabincell{c}{33/22.2 \\ 29/20.4 \\27/19.3}  &\tabincell{c}{29/20.8 \\ 31/21.4 \\34/24.3}  &\tabincell{c}{31/21.2 \\ 33/23.2 \\28/19.9}  &\tabincell{c}{31.2/21.7 \\ 30.8/21.3 \\30.4/21.9}  \\
\midrule
  A3C (V+A+Mapper)    &   \tabincell{l}{Ring \\ Alarm \\ Clock}&  \tabincell{c}{24/17.9 \\ 22/16.6 \\25/18.5}  &\tabincell{c}{30/21.3 \\ 29/20.1 \\31/21.7}  &\tabincell{c}{21/15.6 \\ 23/16.9 \\22/16.2}  &\tabincell{c}{19/14.3 \\ 20/15.2 \\20/14.9}  &\tabincell{c}{24/17.4 \\ 25/17.9 \\24/17.2}  &\tabincell{c}{23.6/17.3 \\ 23.8/17.3 \\24.4/17.7}  \\ 
\midrule
  A3C(V+A+M+Mapper)   &   \tabincell{l}{Ring \\ Alarm \\ Clock}&  \tabincell{c}{30/20.7 \\ 28/19.1 \\31/21.3}  &\tabincell{c}{38/25.7 \\ 39/26.4 \\37/24.9}  &\tabincell{c}{35/23.8 \\ 33/22.5 \\31/22.5}  &\tabincell{c}{33/23.3 \\ 34/23.5 \\36/24.9}  &\tabincell{c}{33/23.8 \\ 36/24.4 \\32/23.5}  &\tabincell{c}{33.8/23.5 \\ 34.0/23.2 \\33.4/23.4}  \\
\midrule
Ours (no Exp.)    &   \tabincell{l}{Ring \\ Alarm \\ Clock}&  \tabincell{c}{69/42.2 \\ 67/41.7 \\70/43.9}  &\tabincell{c}{55/37.4 \\ 53/35.9 \\58/39.8}  &\tabincell{c}{53/35.4 \\ 55/36.8 \\56/37.7}  &\tabincell{c}{56/38.6 \\ 56/37.9 \\54/36.4}  &\tabincell{c}{59/40.3 \\ 58/39.7 \\61/41.6}  &\tabincell{c}{58.4/38.8 \\ 57.8/38.4 \\59.8/40.0}  \\     
\midrule
  Ours (w/ Exp.)   &   \tabincell{l}{Ring \\ Alarm \\ Clock}&  \tabincell{c}{\textbf{78/58.0} \\ \textbf{76/57.0} \\ \textbf{83/60.1}}  &\tabincell{c}{\textbf{66/47.1} \\ \textbf{62/45.5} \\ \textbf{70/56.7}}  &\tabincell{c}{\textbf{62/43.4} \\ \textbf{65/45.7} \\ \textbf{69/48.7}}  &\tabincell{c}{\textbf{68/55.0} \\ \textbf{66/45.1} \\ \textbf{62/51.1}}  &\tabincell{c}{\textbf{74/51.3} \\ \textbf{69/54.4} \\ \textbf{70/55.1}}  &\tabincell{c}{\textbf{69.6/51.0} \\ \textbf{67.6/49.5} \\ \textbf{70.8/54.4} }  \\
\bottomrule
\end{tabular}
\vspace{-2pt}
\caption{Success rates (\%) / Success weighted by Path Length (SPL) of different audio-visual navigation methods. }
\vspace{-8pt}
\label{tab:result}
\end{table*}

\subsection{Baseline}

We consider seven baseline methods for evaluations.

\begin{compactitem}
	\item \textbf{Random Walk}. This is the simplest heuristic  for the navigation task. The agent randomly samples an action from the action space at each time step. 
	
	\item \textbf{Greedy Search (A)}. This is a sound-only baseline. The agent navigates in the environment only using the sound perception module. The agent moves greedily towards the sound source without observing the structure and layout of the apartment. 
	
	\item \textbf{A3C (V)}~\cite{Mnih2016Asynchronous}. Asynchronous advantage actor-critic (A3C) is a state-of-the-art Deep RL based approach for navigation. This is a goal-oriented, vision-only navigation baseline without memory. 
	
	\item \textbf{A3C (V+A)}.  This is a goal-oriented, audio-vision navigation baseline without memory. In our setting, the input of the A3C model is an audio-visual representation of the current state. For a fair comparison, we concatenate the feature representations extracted from the visual and sound perception networks used in our approach. 
	
	\item \textbf{A3C (V+A+Mapper).} In this setting, the input of the A3C model is an audio-visual representation of the current state and the explicit 2D occupancy maps used in our non-exploration based approach. Th top-down occupancy map is concatenated with the audio-visual representation. 
	
	\item \textbf{A3C (V+A+M).}~\cite{savinov2018semi}. Similar to the baseline used in Savinov~\etal\cite{savinov2018semi}, we have also implemented an A3C equipped with LSTM memory as a baseline for our task. First, the agent navigates the environment for the same number of random walking steps in the exploration mode without a given goal. There is no reward for reaching a goal. At test time, we feed the exploration sequence to the LSTM agent and then let it perform goal-directed navigation without resetting the LSTM state. We expect that the LSTM module can implicitly build an internal representation of them.
	
	\item \textbf{\atcvam.} We further combine the explicit and implicit memory to training the A3C.
		
\end{compactitem}


\subsection{Results}

Table~\ref{tab:result} summarizes the results for the three different sounds sources in the five testing apartments. The proposed audio-visual agent performs better than all baselines across all metrics. In the exploration-and-act setting, our model (w/Exp.) remarkably achieves over a 65\% success rate in average for every sound source, almost doubling the best-performing baseline. Figure~\ref{fig:trajectory} presents the results in two testing rooms in more detail, by plotting the trajectory of our approach against that of the best performing baseline (\atcvam). In the non-exploration based setting, our approach (no Exp.) can also achieve nearly 60\% success rate, even outperform the strongest baseline with explorations. These results indicate that our system is better to leverage the layout of the room and find the shortest path to reach the goal. For instance, the agent tends to come to the door when the sound is in another room, and thus efficiently reaches the goal. More qualitative results are available in the supplementary material.

From Table~\ref{tab:result}, it is not surprising to find that the success rates of the random walk--based approaches are very low, because it is very challenging to conduct navigation in our environment without any prior knowledge. Greedy search is slightly better than random search, as the agent can anticipate the location of the goal from the sound. However, the success rates are still below 15\%, because the agent has no visual prior knowledge of the room layout and misunderstands the shortest geometric distance as the shortest path, which does not always hold in our setting. These results demonstrate that it is important to connect visual and sound information for the audio-visual navigation task. 

The performance of the A3C based approaches is significantly weaker than those previously reported in literature~\cite{Mnih2016Asynchronous}. The key reason is that we focus on generalizations---putting agents in previously unseen environments. This is a much more challenging, but also more realistic scenarios.  We also find that the performance gap is still large between our approach and A3C, which uses both implicit spatial memory from exploration and explicit spatial memory during navigation. We speculate that our approach can disentangle the visual perception and sound perception from path planing, thus providing better generalizability in new testing environments. 









\section{Conclusion}
\label{sec:conclusion}

We tackle a novel problem, audio-visual embodied navigation, in a multi-modal virtual environment. The agent can use the internal structured representation of the environment to efficiently navigate to targets in previously unseen environments. Our multi-modal virtual environment consists of complex apartments with a sound module and realistic layouts. We demonstrate that the state-of-the-art deep reinforcement learning approaches meet difficulties in generalization. Our approach generalizes to new targets and new environments with remarkable results.

\vspace{5pt}
{\small \noindent\textbf{Acknowledgement:} This work is in part supported by ONR MURI N00014-16-1-2007, the Center for Brain, Minds, and Machines (CBMM, NSF STC award CCF-1231216), and IBM Research.}

\addtolength{\textheight}{-5.5cm}





\bibliographystyle{IEEEtran}
\bibliography{reference,egbib}

\end{document}